\def\year{2021}\relax
\documentclass[letterpaper]{article} 
\usepackage{aaai21}  
\usepackage{times}  
\usepackage{helvet} 
\usepackage{courier}  
\usepackage[hyphens]{url}  
\usepackage{graphicx} 
\usepackage{natbib}  
\usepackage{caption} 
\usepackage{amsmath,amssymb,subfigure}
\usepackage{multirow,multicol}
\usepackage[ruled]{algorithm2e}

\usepackage[hidelinks]{hyperref}

\title{Interpretable Embedding Procedure Knowledge Transfer via Stacked Principal Component Analysis and Graph Neural Network}
\author{Seunghyun Lee, Byung Cheol Song\\
}
\affiliations{Department of Electronic Engineering, Inha University, Incheon 22212, South Korea\\
lsh910703@gmail.com, bcsong@inha.ac.kr
}
\begin{document}
\maketitle

\begin{abstract}
    Knowledge distillation (KD) is one of the most useful techniques for light-weight neural networks. Although neural networks have a clear purpose of embedding datasets into the low-dimensional space, the existing knowledge was quite far from this purpose and provided only limited information. We argue that good knowledge should be able to interpret the embedding procedure. This paper proposes a method of generating interpretable embedding procedure (IEP) knowledge based on principal component analysis, and distilling it based on a message passing neural network. Experimental results show that the student network trained by the proposed KD method improves 2.28\% in the CIFAR100 dataset, which is higher performance than the state-of-the-art (SOTA) method. We also demonstrate that the embedding procedure knowledge is interpretable via visualization of the proposed KD process.
    The implemented code is available at \url{https://github.com/sseung0703/IEPKT}.
\end{abstract}

\section{Introduction}\label{sec.1}
    Convolutional neural networks (CNNs) have been adopted by a variety of areas because of their outstanding performance. However, CNNs require a huge amount of computation and memory cost, which makes it hard to mount on embedded and mobile systems. Knowledge distillation (KD) is one of the solutions to build light-weighted CNNs~\cite{KD}. The main function of KD is to create and deliver a certain knowledge so that a student network behaves similarly to a teacher network. Since KD can be applied to various machine learning areas such as semi-supervised learning and zero-shot learning, KD has been received a lot of attention recently. Conventional KD algorithms defined information from several locations of CNN, e.g., intermediate feature maps~\cite{FitNet,kd-attention} and embedded feature vectors at the output end of CNN~\cite{KD,rkd,kd-ft}, as the knowledge of CNN.
    
    Note that CNN's ultimate goal is to map high dimensional data such as images and audio to low dimensional space for easy analysis. However, the knowledge proposed so far has been far from the purpose of CNNs. In order to improve the student network's embedding performance, it is important to accurately convey the information about the embedding process of CNN, which analyzes a dataset in order from low-level features to high-level features. Therefore, we insist that the knowledge of CNN should represent the embedding procedure and be able to interpret human insight during the distillation process. In order to define and distill such interpretable embedding procedure (IEP) knowledge, we have gained a few insights from the following previous works.

    \begin{figure}[t]\centering
        \includegraphics[width=0.35\textwidth]{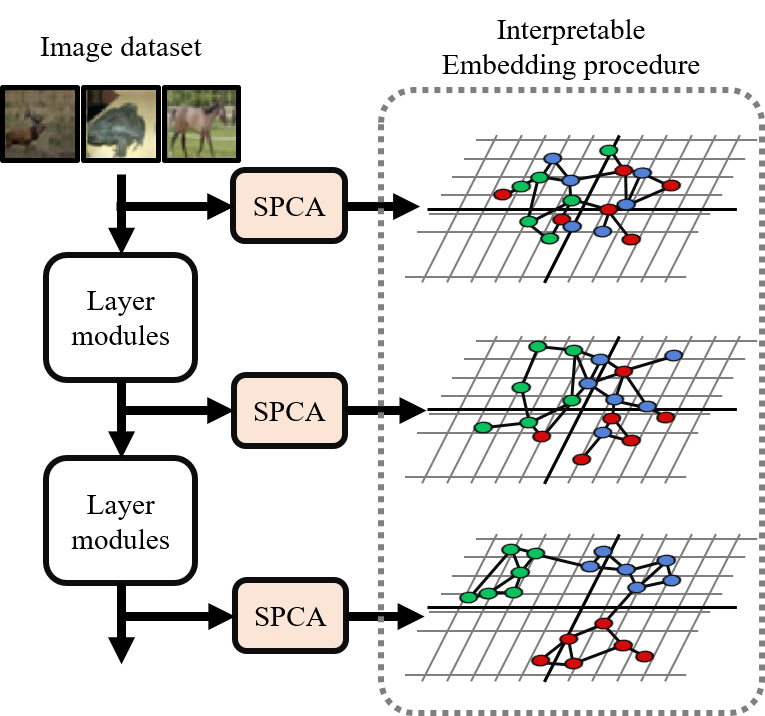}
        \caption{The conceptual illustration of the neural network's embedding procedure knowledge with the proposed stacked principal component analysis (SPCA).}\label{fig1}
    \end{figure}

    Principal component analysis (PCA) has been one of the effective tools for visualizing and analyzing embedded feature distribution~\cite{gan_pca}. On the other hand, since graph-structure can effectively represent inter-data relations, many graph neural networks (GNNs) have been developed recently. For example, GNN is attracting attention as a reliable solution to represent embedding space~\cite{mpnn_emb}.

    Based on insights from the previous works, this paper presents a new KD method for distilling IEP knowledge. First, to analyze the embedding procedure of the teacher network, a stacked PCA (SPCA), which performs PCA twice, is proposed. SPCA allows the feature map dimension to be shallow, enabling analysis and visualization of embedding procedures at relatively low cost. Here, the graph-structure is employed. Figure~\ref{fig1} illustrates this concept. Next, in order to distill IEP knowledge with minimal information loss, a new distillation method using a message passing neural network (MPNN) is proposed. The MPNN distills IEP knowledge by estimating the embedding procedure of each stage from the previous stage's graph.

    Our contribution points are as follows: First, student networks trained by the proposed IEP knowledge provide SOTA performance. Second, IEP knowledge is interpretable via visualization, which represents the embedding procedure of CNN and coincides with human insight.
    
\section{Related Works}\label{sec.2}
    \subsection{Knowledge Distillation}\label{sec.2.1}
        Conventional KD methods defined various knowledge. Some of them defined the neural response of the last or several interim feature maps of CNN as knowledge~\cite{kd-vid,KD,FitNet,kd-attention}. Also, there have been attempts to distill embedding knowledge so as to overcome the problem of lack of knowledge in existing KD techniques~\cite{selective-KD,mhgd,rkd}. Such approaches were mainly intended to find inter-data relations in the embedding space of the last stage of CNN. Only some of the approaches distilled embedding procedure information directly~\cite{mhgd}. However, the knowledge defined by the above-mentioned methods was not interpretable, which means that the previous methods could lose significant information during the distillation process. To distill the teacher network's knowledge with no or minimal information loss, we propose a new method for distilling IEP knowledge.
        
    \subsection{PCA in Deep Learning}\label{sec.2.2}
        As derivative functions of singular value decomposition (SVD) and eigendecomposition (EID) have been defined recently~\cite{svd_grad}, several studies to fuse SVD and EID with deep neural networks have been published~\cite{KD-SVD,svd_grad_example}. For example, SVD was used to compress feature maps~\cite{svd_grad_example}, and principal components themselves were employed as compressed feature vectors~\cite{mhgd,KD-SVD}. As one of the ways to reduce the PCA's memory cost, incremental PCA (IPCA) incrementally estimated dataset's principal components and predicted mean vectors on a mini-batch basis~\cite{ipca}. Recently, applying PCA to deep learning have been proposed~\cite{ipca_example}. From the previous studies, we got an insight that embedding procedure knowledge can be obtained from the teacher network through IPCA.
        
    \subsection{Graph Neural Network}\label{sec.2.3}
        GNN has been studied in various fields as a tool for obtaining inter-data relations. In particular, MPNN, one of the GNNs, has emerged as a core technology in areas where edge information is important, such as physics~\cite{physics1,physics0} and chemistry tasks~\cite{mpnn,chemical1}. Recently, Meng et al. proposed a technique to apply MPNN to embedding task~\cite{mpnn_emb}. The method was able to interpret and estimate complex relationships between data by defining inter-data relations as edges. From the previous works, we gained another insight that MPNNs can capture relational information better than the attention networks that become recently popular more and more.

\section{Methods}\label{sec.3}
    \begin{figure}[t]\centering
        \includegraphics[width=0.45\textwidth]{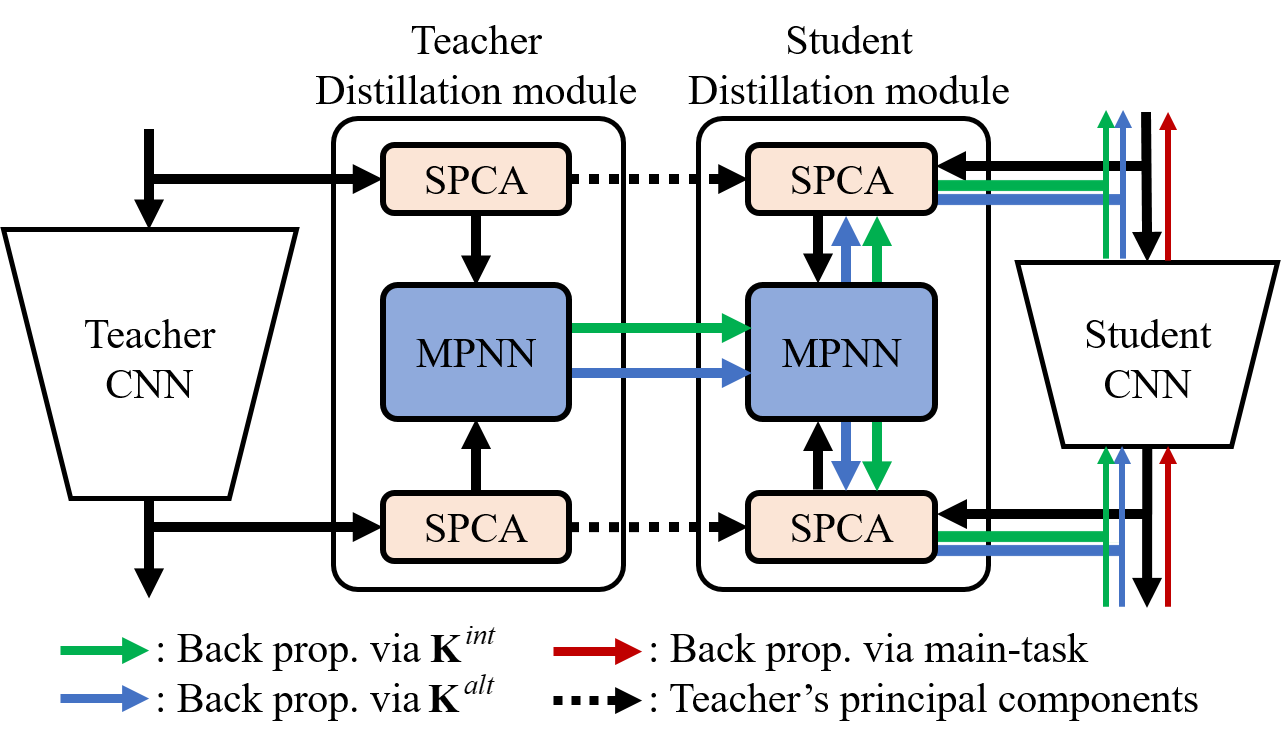}
        \caption{Conceptual diagram of the proposed knowledge distillation algorithm. $\textbf{K}^{int}$ and $\textbf{K}^{alt}$ mean knowledge of interim embedding stages and alteration of them, respectively.}\label{fig2}
    \end{figure}
    This section defines knowledge of embedding procedure, i.e., the goal of neural networks, and suggests how to distill the knowledge in an interpretable form. The block diagram of the proposed method is shown in Fig.~\ref{fig2}. First, compress the set of feature maps of the teacher network with SPCA, and calculate the embedding procedure, and define it as knowledge. MPNN distills this knowledge in a form that can be transferred to the student network. Next, the student network's IEP knowledge is distilled through a distillation module trained with the teacher network in advance. Finally, the student network's target task and teacher's knowledge are trained via multi-task learning. In addition, this paper presents a method for integrating the proposed method with multi-head graph distillation (MHGD), one of the latest KD techniques, to accomplish complete neural network knowledge.

    \begin{figure}[t]\centering
        \includegraphics[width=0.4\textwidth]{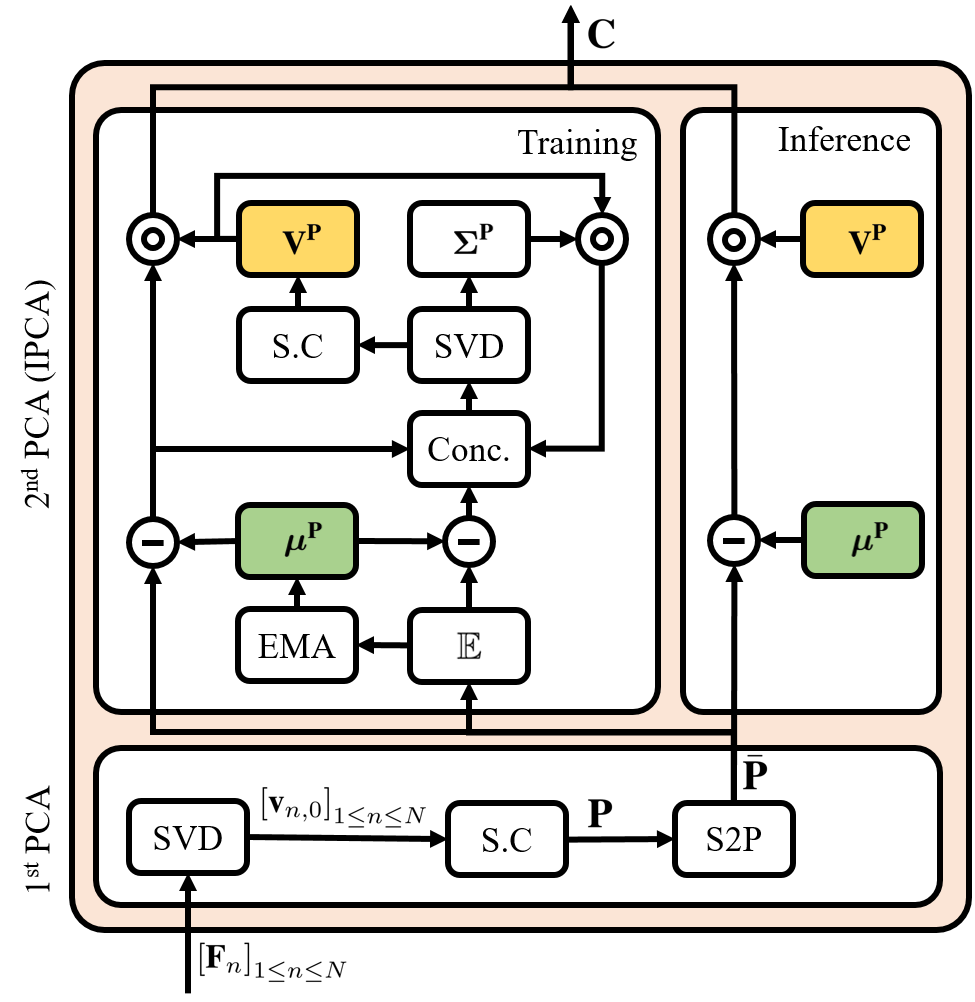}
        \caption{The block diagram of the stacked PCA (SPCA). Here, S.C is sign correction, S2P is sphere to plane mapping, EMA is an exponential moving average, and conc. means concatenation.}\label{fig3}
    \end{figure}

    \subsection{Producing IEP Knowledge via SPCA}\label{sec.3.1}
        In order to produce embedding procedure knowledge of CNN, we need to analyze the inter-data relation of feature maps sensed in the middle of CNN, which corresponds to the intermediate stage of embedding. However, finding relations between feature maps usually requires high computation cost. Therefore, SPCA is applied to effectively reduce the dimension of feature maps without crucial information loss. SPCA also contributes to visualizing a set of feature maps for better understanding. The conceptual diagram of the SPCA is shown in Fig.~\ref{fig3}.

        Since feature maps generally have high spatial correlation, some feature vectors can have similar information to each other. Thus, a feature map can be approximated with several principal components (PCs). In detail, a feature map of \textit{H}$\times$\textit{W}$\times$\textit{D} is converted into a matrix \textbf{F} of \textit{HW}$\times$\textit{D} by PCA. Here, \textit{H}, \textit{W}, \textit{D} indicate the width, height, and depth of a feature map, respectively. We only adopt the first PC to compress the feature map as much as possible. Next, the sign flipping, which makes the largest absolute value positive, is used to remove the sign ambiguity of singular vectors (S.C in Fig.~\ref{fig3}). Finally, assuming a mini-batch data, a set of PCs \textbf{P} is represented as follows.
        \begin{equation}\label{eq1}
            \textbf{P}=\left[\textbf{p}_{n}\right]_{1\leq n \leq N}=\left[s_{n}\textbf{v}_{n,0}\right]_{1\leq n \leq N},\ \textbf{P}\in \mathbb{R}^{N\times D}
        \end{equation}
        \begin{equation}\label{eq2}
            s_{n}=\text{sign}\left(\text{max}\left(\textbf{v}_{n,0}\right)+\text{min}\left(\textbf{v}_{n,0}\right)\right)
        \end{equation}
        \begin{equation}\label{eq3}
            \textbf{V}_{n}=\left[\textbf{v}_{n,k}\right]_{1\leq k \leq \text{min}\left(HW,D\right)},\ \textbf{F}_{n}=\textbf{U}_{n}\boldsymbol{\mathbf{\Sigma}}_{n}\textbf{V}^{*}_{n}
        \end{equation}
        where * is the Hermitian function and \textit{N} is the batch size.

        Since $\textbf{p}_{n}$ is \textit{HW} times smaller than $\textbf{F}_{n}$, it is possible to obtain the relation with very low cost. But the dimension of \textbf{P} is still too high to be interpreted by humans. So we apply PCA once more to find the PCs of the embedding space (the $2^{\text{nd}}$ PCA in Fig. \ref{fig3}). Additional dimension reduction by the second PCA not only enables visualization but also eliminates unnecessary information. However, since $\textbf{p}_{n}$ exists on the hypersphere as a unit vector, it is difficult to obtain a linear least squares solution. Therefore, as shown in Eq.~(\ref{eq4}), we apply stereographic projection \cite{stereographic} for mapping $\textbf{P}$ to a plane space.
        \begin{equation}\label{eq4}
            \bar{\textbf{P}}=\left[\bar{\textbf{p}}_{n}\right]_{1\leq n \leq N}=\left[\frac{\textbf{p}_{n}+\textbf{o}}{\text{cos}\left(\text{cos}^{-1}\left(\textbf{p}^{*}_{n}\cdot\textbf{o}\right)/2\right)^{2}}-2\textbf{o}\right]_{1\leq n \leq N},
        \end{equation}
        where \textbf{o} is the center vector of the space where PCs exist, that is $\left[1/\sqrt{D},...,1/\sqrt{D} \right]$.
        
        Now, since $\bar{\textbf{P}}$ is on the plane space, PCA can be applied. However, only mini-batch data is available for learning, and if the batch size is smaller than the feature dimension, full-rank PCs cannot be obtained. Also, it is not reasonable to assume that the PCs of the mini-batch data match those of the dataset. To solve these problems, we employ IPCA~\cite{ipca} which produces approximated PCs $\textbf{V}^{\bar{\textbf{P}}}$ and mean vector $\boldsymbol{\mu}^{\bar{\textbf{P}}}$ by iteratively updating them. The IPCA in Fig.~\ref{fig3} is identical to the conventional IPCA, except that the mean vector calculation is replaced with an exponential moving average (EMA). The detailed description is given in the supplementary material. Since only the top half of the total PCs are used, the dimension of a set of compressed PCs \textbf{C} is reduced to half than before, as in Eq.~(\ref{eq5}) and Eq.~(\ref{eq6}).
        \begin{equation}\label{eq5}
            \textbf{C}=\left[\textbf{c}_{n}\right]_{1\leq n\leq N}=\left[\left(\bar{\textbf{p}}_{n}-\boldsymbol{\mu}^{\bar{\textbf{P}}}\right)\cdot \textbf{V}^{\bar{\textbf{P}}}\right]_{1\leq n\leq N}\boldsymbol{\Sigma}^{\bar{\textbf{p}}}
        \end{equation}
        \begin{equation}\label{eq6}
            \textbf{C}\in \mathbb{R}^{N \times D/2}
        \end{equation}
        Finally, based on Eq. (\ref{eq5}) and definition of cosine similarity, an affinity matrix $\textbf{A}$ is defined by
        \begin{equation}\label{eq7}
            \textbf{A}=\left[\frac{1}{\left\|\textbf{c}_{v}\right\|_{2}\left\|\textbf{c}_{w}\right\|_{2}}\textbf{c}^{*}_{v}\cdot \textbf{c}_{w} \right]_{1\leq v,w \leq N}
        \end{equation}
        Note that $\textbf{A}$ has intermediate embedding information at a sensing point of CNN. Assuming that feature maps sensed in $L$ points, IEP knowledge is obtained by observing a set of intermediate embedding informations $\left[\textbf{A}_l\right]_{1 \leq l \leq L}$ and a set of their alterations $\left[\textbf{A}_{l+1}-\textbf{A}_l\right]_{1 \leq l \leq L-1}$. Also, when we extract the top three components in \textbf{C} for three-dimensional visualization, it can be easily understood through visualization (see the supplementary material). The next section describes how to distill IEP knowledge in a transferable form.
        
    \subsection{MPNN for Distilling IEP Knowledge}\label{sec.3.2}
\begin{figure}[t]\centering
            \includegraphics[width=0.4\textwidth]{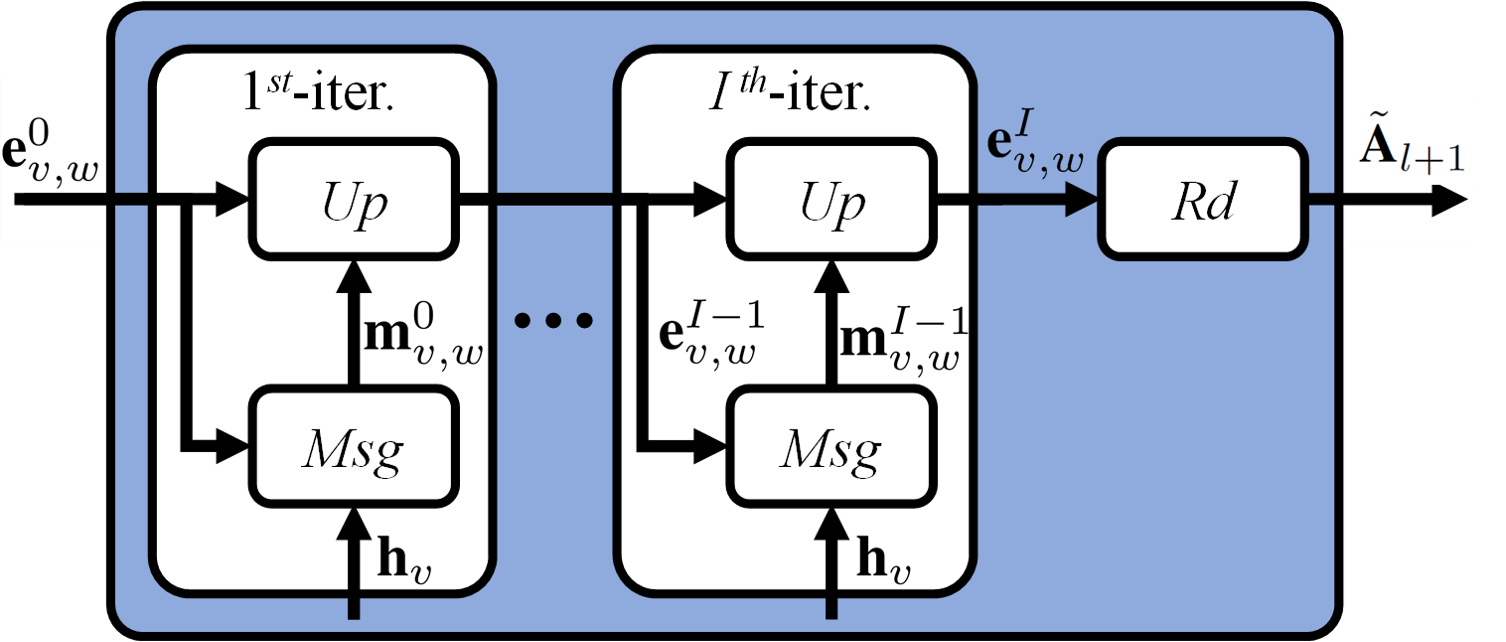}
            \caption{The block diagram of the proposed MPNN to distill interpretable embedding procedure knowledge. }\label{fig4}
        \end{figure}

        The IEP knowledge obtained by the SPCA coincides with the purpose of CNN. But in the case of a simple student network, this knowledge is so sharp and complex, which can give over-constraint if transferred as is. So, in order to distill the IEP knowledge, we employ MPNN which can interpret inter-data relation effectively and give a task that estimates the next interim embedding stages from current stages. The overall structure of the proposed MPNN is illustrated in Fig.~\ref{fig4}.
        
        First, the node feature $\textbf{h}_{v}$ and the initial edge feature $\textbf{e}_{v,w}^{0}$ are defined using the two set of compressed PCs $\textbf{C}_{l}$ and $\textbf{C}_{l+1}$ which are sensed at two adjacent points. $\textbf{h}_{v}$ uses $\textbf{c}_{l+1,v}$ as it is, and $\textbf{e}_{v,w}^{0}$ is obtaining by linear mapping (\textit{LM}) of $\textbf{c}_{l,v}$ and dimension-wise relation like Eq.~(\ref{eq8}).
        \begin{equation}\label{eq8}
            \textbf{e}^{0}_{v,w}=\bar{\textbf{c}}_{l,v}\odot \bar{\textbf{c}}_{l,w},\ \bar{\textbf{c}}_{l,v}=\frac{\textit{LM}\left(\textbf{c}_{l,v}\right)}{\left\|\textit{LM}\left(\textbf{c}_{l,v}\right)\right\|_{2}},
        \end{equation}
        where $\odot$ stands for Hadamard product, and \textit{LM} consists of a fully connected (FC) layer and a batch normalization~\cite{batchnorm}. Next, to update the edge feature, the message function \textit{Msg} and the edge update function \textit{Up} operate as follows.
        \begin{equation}\label{eq9}
            \textbf{m}^{i}_{v,w}=\textit{Msg}\left(\textbf{h}_{v},\textbf{h}_{w},\textbf{e}^{i}_{v,w}\right)                =\textit{GLU}\left(\left[\textbf{h}_{v}-\textbf{h}_{w},\mathbb{E}\left(\textbf{e}^{i}_{v,w}\right) \right]\right)
        \end{equation}
        \begin{equation}\label{eq10}
            \textbf{e}^{i+1}_{v,w}=\textit{Up}\left(\textbf{e}^{i}_{v,w},\textbf{m}^{i}_{v,w}\right)=\textbf{e}^{i}_{v,w}+\textbf{m}^{i}_{v,w},
        \end{equation}
        where $\mathbb{E}$ is a function that returns the average of all components of an input. \textit{Msg} uses a gated linear unit (GLU) \cite{glu}, and \textit{Up} simply adds edge features and messages. Finally, the edge feature $\textbf{e}^{\textit{I}}_{v,w}$ is obtained by repeating the above process \textit{I} times, and it is input to the readout function \textit{Rd}, and the (\textit{l} + 1)$^{\textit{th}}$ affinity matrix $\tilde{\textbf{A}}_{l+1}$ is returned. This final process is expressed as follows.
        \begin{equation}\label{eq11}
            \textit{Rd}\left(\textbf{e}^{I}_{v,w}\right)=\mathbb{E}\left(\textbf{e}^{I}_{v,w}\right)
        \end{equation}
        \begin{equation}\label{eq12}
            \tilde{\textbf{A}}_{l+1}=\left[\textit{Rd}\left(\textbf{e}^{I}_{v,w}\right)\right]_{1\leq v,w \leq N}
        \end{equation}
        We use Kullback-Leibler divergence (KLD)~\cite{kld} as a loss function for learning the proposed MPNN.
        \begin{equation}\label{eq13}
            \mathcal{L}^{\textit{MPNN}}=\textit{KLD}\left(\sigma\left(\textbf{A}_{l+1}\right) \| \sigma\left(\tilde{\textbf{A}}_{l+1} \right) \right),
        \end{equation}
        where $\sigma$ stands for the softmax function. The details of learning are described in the supplementary material.
        
        In the proposed MPNN, each edge and message have clear meanings. The initial edge indicates the \textit{l}$^{\textit{th}}$ interim embedding stage, and MPNN updates it to the (\textit{l}+1)$^{\textit{th}}$ interim embedding stage using messages, which indicate alteration of embedding stage. Therefore, we define them as an intermediate embedding knowledge $\textbf{K}^{\textit{int}}$ and alteration of embedding knowledge $\textbf{K}^{\textit{alt}}$ to transfer IEP knowledge, which are shown in Eq.~(\ref{eq14}) and Eq. (\ref{eq15}), respectively.
        \begin{equation}\label{eq14}
            \textbf{K}^{\textit{int}}=\left[\tilde{\textbf{A}}_{l+1}\right]_{1 \leq l \leq L-1}
        \end{equation}
        \begin{equation}\label{eq15}
            \textbf{K}^{\textit{alt}}=\left[\textbf{m}^{i}_{l,v,w}\right]_{1\leq l \leq L-1,\ 1\leq v,w \leq N,\ 1\leq i \leq I}
        \end{equation}
        
        Since this knowledge is smoothened by neural layers, it can be more easily learned by the student network. Also, the proposed knowledge can interpret the embedding procedure through visualization, which can be verified in the Experiment Section.
        
    \subsection{Transferring IEP Knowledge to the Student}\label{sec.3.3}
        To transfer the IEP knowledge distilled from the teacher network, a distillation module must be applied even to the student network. First, SPCA is applied to the matrix-formed feature map $\textbf{F}^{S}_{l}$ which is sensed in the student network. Next, $\textbf{C}^{S}_{l}$ is obtained using the information produced by the SPCA of the teacher network as follows.
        \begin{equation}\label{eq16}
            \textbf{P}^{S}_{l}=\left[\textbf{p}^{S}_{l,n}\right]_{1\leq n \leq N}=\left[\frac{s_{n}\left(\textbf{u}^{S}_{l,n}\right)^{*}\cdot \textbf{F}^{S}_{n}}{\left\| \left(\textbf{u}^{S}_{l,n}\right)^{*}\cdot \textbf{F}^{S}_{n}\right\|_{2}}\right]_{1\leq n \leq N}
        \end{equation}
        \begin{equation}\label{eq17}
            \textbf{C}^{S}_{l}=\left[\left(\textbf{p}^{S}_{l,n}-\boldsymbol{\mu}^{\textbf{P}}\right)\cdot \textbf{V}^{\textbf{P}} \right]_{1 \leq n \leq N}
        \end{equation}
        The generated $\textbf{C}^{S}_{l}$ is then inputted into the MPNN, which shares parameters with the teacher network, to distill the knowledge of the student network. The knowledge should be transferred in a way that minimizes the difference between the distilled knowledge of the teacher and student networks. So, the loss functions $\mathcal{L}^{\textit{int}}$ and $\mathcal{L}^{\textit{alt}}$ are defined by Eq. (\ref{eq18}) and (\ref{eq18_}).
        \begin{equation}\label{eq18}
            \mathcal{L}^{\textit{int}}=\textit{KLD}\left(\sigma\left(\textbf{K}^{S,\textit{int}}\right)\|\sigma\left(\textbf{K}^{T,\textit{int}}\right)\right)
        \end{equation}
        \begin{equation}\label{eq18_}
            \mathcal{L}^{\textit{alt}}=\frac{1}{N^{2}}\left\|\textbf{K}^{S,\textit{alt}}-\textbf{K}^{T,\textit{alt}} \right\|_{1}
        \end{equation}
        KLD is adopted to avoid putting too strong constraints on $\textbf{K}^{int}$ (the 2$^{\textit{nd}}$ term of Eq. (\ref{eq18})). Since $\textbf{K}^{alt}$ is a key information, strong constraint using $L_{1}$-norm is given to $\textbf{K}^{alt}$ (the 1$^{\textit{st}}$ term of Eq. (\ref{eq18})). So, the proposed method transfers two kinds of knowledge to the student network. Thus, the student network learns totally three tasks simultaneously, including a target task. 
        
        However, when transferring knowledge in the middle of CNN, the gradient of the transfer task can be much larger than that of the target task. This becomes over-constraint on the student network. Therefore, an appropriate multi-task learning technique is required to prevent such a phenomenon.

        \cite{KD-SVD} proposed a gradient clipping based on the norm of the target task's gradient. Inspired from \cite{KD-SVD}, we propose to clip the gradient obtained by knowledge based on the norm of the target task's gradient. Specific details are as follows.
        \begin{equation}\label{eq19}
            \frac{\partial \Theta}{\partial \mathcal{L}^{\textit{Total}}}=\frac{\partial \Theta}{\partial \mathcal{L}^{\textit{Target}}}+\textit{clip}\left(\frac{\partial \Theta}{\partial \mathcal{L}^{\textit{int}}}\right)+\textit{clip}\left(\frac{\partial \Theta}{\partial \mathcal{L}^{\textit{alt}}}\right)
        \end{equation}
        \begin{equation}\label{eq20}
            \textit{clip}\left(z\right)=\text{max}\left(1,\left\|\frac{\partial \Theta}{\partial \mathcal{L}^{\textit{Target}}}\right\|_{2}/\left\|z\right\|_{2}\right)z
        \end{equation}
        As a result, the knowledge of the teacher network can be transferred as much as possible without over-constraint.
        
    \begin{figure*}[t]\centering
        \includegraphics[width=0.9\textwidth]{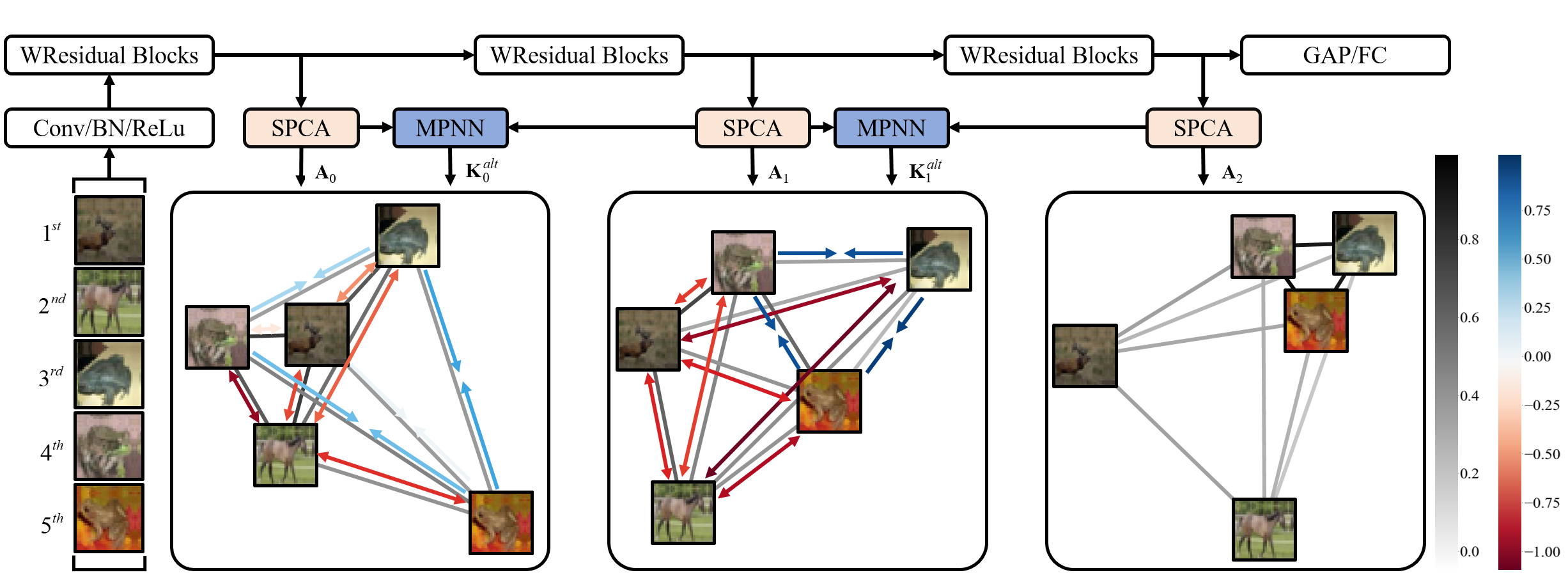}
        \caption{Visualization of $\textbf{A}$ and $\textbf{K}^{alt}$ obtained through the proposed distillation module. $\textbf{A}$ and $\textbf{K}^{alt}$ contain information of interim embedding stages and their alterations, respectively. For clear visualization, $\textbf{A}$ is represented by gray scale and $\textbf{K}^{alt}$ by RdBu colormap.}\label{fig5}
    \end{figure*}
        
    \subsection{Black-box Knowledge Distillation via Multi-head Graph Distillation}\label{sec.3.4}
        The proposed method focuses on distilling CNN's IEP knowledge. However, in fact, CNN has black-box knowledge that cannot be interpreted due to its inherent characteristic. So distilling black-box knowledge is required to deliver complete knowledge of CNN. We generate black-box knowledge $\textbf{K}^{BB}$ by fusing the multi-head graph distillation (MHGD)~\cite{mhgd} and linguistic-informed self-attention (LISA)~\cite{LISA}. MHGD is a method of distillation of arbitrary relations interpreted by CNN through an attention network. LISA also adds extrinsic information to the attention network so that the attention network can extract an arbitrary relation of data. Next, similarly to LISA, the black-box knowledge $\textbf{K}^{BB}$ is distilled by injecting the IEP knowledge into MHGD. The detailed structure is explained in the supplementary material.

        The proposed method can transfer all kinds of embedding procedure knowledge from the teacher network, i.e., black-box knowledge as well as IEP knowledge, to the student network. Therefore, the student network receives clear guidance about the embedding procedure from the teacher network, which leads to additional performance improvement.
        
\section{Experimental Results}\label{sec.4}
    This section shows the results of three kinds of experiments. First, we visualize the information of the proposed IEP knowledge. Second, we evaluate the basic performance of the proposed method through experiments on small network enhancement and transfer learning. Third, the effect of each component of the proposed method on the overall performance is verified. Additional evaluation results to show the effectiveness of the proposed method can be found in the supplementary material. We used four neural network architectures for implementing the proposed method: WResNet~\cite{wideresnet}, ResNet~\cite{resnet}, MobileNet-V2~\cite{mobilenetv2}, and VGG~\cite{vgg}. Also, one of the most popular methods, attention transfer (AT)~\cite{kd-attention} as well as four SOTA methods: factor transfer (FT)~\cite{kd-ft}, activation boundary (AB)~\cite{AB-KD}, relational knowledge distillation (RKD)~\cite{rkd}, and comprehensive overhaul (CO)~\cite{heo2019comprehensive}. were adopted for comparison with the proposed method. In addition, MHGD~\cite{mhgd}, which has a similar concept to the proposed method, was compared. We implemented all the techniques to be compared by ourselves. Detailed information such as network structures and hyper-parameters is described in the supplementary material.
    
    \begin{table*}[t]
        \begin{center}
        \begin{tabular}{c|c|c|cccccc|cc}
        \hline\hline
        Dataset & Rate & Student & AT & FT & AB & RKD & MHGD & CO & IEP & IEP+Black-box\\
        \hline
        \hline
        \multirow{4}{*}{CIFAR100} & Full & 76.09 & 76.98 & 77.14 & 77.29 & 77.02 & 77.45 & 78.21 & 78.12 & \textbf{78.37}\\
                                  & 0.50 & 69.77 & 71.13 & 72.41 & 72.28 & 69.57 & 73.32 & 74.33 & 74.22 & \textbf{74.53}\\
                                  & 0.25 & 59.28 & 63.07 & 63.70 & 66.79 & 53.57 & 67.27 & 67.90 & 68.57 & \textbf{69.02}\\
                                  & 0.10 & 40.65 & 47.66 & 48.29 & 57.38 & 23.27 & 54.58 & 40.80 & 55.89 & \textbf{59.04}\\
        \hline
        \multirow{4}{*}{TinyImageNet} & Full & 59.71 & 60.92 & 55.61 & 60.19 & 61.12 & 62.26 & 63.56 & 63.29 & \textbf{63.73}\\
                                      & 0.50 & 52.53 & 54.50 & 55.81 & 54.41 & 54.09 & 56.56 & 59.14 & 58.56 & \textbf{59.27}\\
                                      & 0.25 & 43.56 & 46.54 & 39.19 & 48.99 & 42.19 & 50.59 & 52.56 & 53.20 & \textbf{53.68}\\
                                      & 0.10 & 28.44 & 32.38 & 34.08 & 42.18 & 20.90 & 38.28 & 34.73 & 43.00 & \textbf{45.01}\\
        \hline
        \end{tabular}
        \caption{Small network enhancement performance comparison of several KD methods for CIFAR100 and TinyImageNet datasets with various sample rates.}\smallskip
        \label{Table1}
        \end{center}
    \end{table*}
    
    \begin{table*}[t]
        \begin{center}
        \begin{tabular}{c|c|c|cccccc|cc}
        \hline\hline
        Dataset & Rate & Student & AT & FT & AB & RKD & MHGD & CO & IEP & IEP+Black-box\\
        \hline
        \hline
        \multirow{4}{*}{CUB200-2011} & Full & 52.21 & 58.87 & 59.96 & 56.80 & 52.54 & 55.77 & 60.83 & 60.13 & \textbf{61.35}\\
                                     & 0.50 & 30.58 & 39.51 & 42.94 & 39.77 & 29.72 & 34.02 & 37.61 & 42.24 & \textbf{43.06}\\
                                     & 0.25 & 14.25 & 19.68 & 21.18 & 20.52 & 14.15 & 18.41 & 14.29 & 22.00 & \textbf{22.60}\\
                                     & 0.10 & 5.87 & 8.05 & 8.04 & 7.03 & 6.60 & 5.97 & 4.61 & 8.74 & \textbf{9.69}\\
        \hline
        \multirow{4}{*}{MIT-scene}   & Full & 51.00 & 56.32 & 60.07 & 59.52 & 53.50 & 47.90 & 57.72 & 59.32 & \textbf{60.94}\\
                                     & 0.50 & 36.83 & 42.43 & 46.53 & 46.80 & 39.18 & 36.48 & 35.16 & 45.83 & \textbf{47.85}\\
                                     & 0.25 & 21.59 & 28.54 & 31.96 & 33.13 & 25.39 & 25.51 & 21.14 & 33.83 & \textbf{34.28}\\
                                     & 0.10 & 10.59 & 14.44 & 14.39 & 19.79 & 12.17 & 10.07 &  6.07 & 18.44 & \textbf{19.94}\\
        \hline
        \end{tabular}
        \caption{Transfer learning performance comparison of several KD methods for CUB-200-2011 and MIT-scene datasets with various sample rates.}\smallskip
        \label{Table2}
        \end{center}
    \end{table*}
    
    \subsection{Visualization of Embedding Procedure}\label{sec.4.1}
        This section visualizes the proposed IEP knowledge that is configured by $\textbf{A}$ and $\textbf{K}^{\textit{alt}}$. For this experiment, WResNet40-4 trained on CIFAR10 dataset~\cite{cifar} was used. First, color mapping was employed to clearly display the numerical values of the each knowledge. Next, the edge lengths were manually adjusted according to the strength of the relation. Figure 5 shows the visualization result.

        Note that through the neural layers of CNN, a given data is embedded as it is analyzed from low-level features to high-level features. If the embedding procedure of CNN is visualized, it is expected that data can be clustered according to the similarity of low-level features in the early stage of CNN and the similarity of high-level features in the late stage. In fact, Fig.~\ref{fig5} shows that the changes in the affinity matrices of the proposed knowledge are consistent with the above expectation. The feature maps of the initial layer have a strong relation to visually similar data, i.e., the 1$^{\textit{st}}$ to 4$^{\textit{th}}$ feature maps. On the other hand, since high-level features are analyzed as the later layers progress, we can observe that feature maps has a strong relation to the context, that is, the data of the same class, e.g., the 3$^{\textit{rd}}$, 4$^{\textit{th}}$, and 5$^{\textit{th}}$ feature maps. Accordingly, $\textbf{K}^{\textit{alt}}$ obtained by the proposed method has a positive value for data of the same class and a negative value for data of different classes.
        
        Thus, the proposed IEP knowledge can be a tool for describing the embedding procedure, which is consistent with human intuition. Additional visualization results about full data in CIFAR10  can be found in the supplementary material.
    
    \subsection{Small Network Enhancement with Sampled Dataset}\label{sec.4.2}
        This section examines the small network enhancement performance of the proposed method on CIFAR100~\cite{cifar} and TinyImageNet~\cite{tinyimagenet}. In this experiment, the teacher and student networks were trained on the same dataset, but the student network used sampled datasets. Four sampling rates (full, 0.5, 0.25, and 0.1) were considered. WResNet40-4 was used as the teacher network. The teacher network provides the performance of 77.52\% on CIFAR100 and 62.30\% on TinyImageNet. We used WResNet16-4 as the student network, and set the hyper-parameters such that we can get the best performance when using the full dataset (rate = full).
        
        Since the proposed method transfers embedding procedure knowledge, high performance can be expected thanks to the excellence of knowledge if the teacher and student networks are trained with the same dataset. Table~\ref{Table1} proves this assumption by comparing the proposed method with the existing KD methods. We can find that the proposed method outperforms other KD schemes at all sampling rates. The larger the sampling rate, that is, the lower the rate, the better the proposed method. For example, when the rate is 10\%, the IEP+Black-box at CIFAR100 performed 1.66\% higher than AB that is the best of SOTA techniques. Also, it provided 2.83\% higher performance than AB for TinyImageNet. This is because the proposed KD method can transfer the most accurate knowledge that the student network needs to perform the target task.
    
    \subsection{Transfer Learning}\label{sec.4.3}
        The next experiment is about transfer learning. As the teacher network, ResNet32 which was pre-trained with ImageNet-2012~\cite{imagenet} was employed. As the student network, ResNet14 which was learned with MIT-scene~\cite{mit-scene} and CUB200-2012~\cite{cub200} datasets was used. Also, the student network was pre-trained with teacher knowledge and fine-tuned with the target dataset. The experimental results are shown in Table~\ref{Table2}. We can observe that the proposed method, IEP+Black-box, is always superior to conventional techniques. However, because the target datasets of the teacher and student networks do not match, the performance improvement in Table~\ref{Table2} tends to be somewhat lower than that in Table~\ref{Table1}. Therefore, to apply the proposed method most effectively, it is important to use the teacher network trained on the same dataset as the target dataset.
    
    \begin{table*}[t]
        \begin{center}
        \begin{tabular}{c|c|ccccccc}
        \hline\hline
        Architecture & Student & AT & FT & AB & RKD & MHGD & CO & IEP+Black-box\\
        \hline
        \hline
        WResNet16-2 & 56.61 & 59.42 & 57.28 & 62.53 & 54.27 & 59.29 & 60.21 & \textbf{63.78}\\
        WResNet16-1 & 51.88 & 53.01 & 50.95 & 55.01 & 48.46 & 50.72 & 52.67 & \textbf{56.09}\\
        MobileNet-V2 & 56.96 & 59.04 & 57.48 & 61.35 & 58.17 & 61.80 & 62.72& \textbf{64.82}\\
        VGG & 47.76 & 49.88 & 48.13 & N/A & N/A & 47.40 & 45.18 & \textbf{55.82}\\

        \hline
        \end{tabular}
        \caption{Small network enhancement performance comparison for different architecture or feature depth with teacher network.}\smallskip
        \label{Table3}
        \end{center}
    \end{table*}
    \subsection{Knowledge Transfer to Heterogeneous Student Network}\label{sec.4.4}
        In this section, we tried to transfer teacher knowledge to several heterogeneous student networks that have different architectures from the teacher network to verify knowledge's dependency on network architecture. As a dataset for this experiment, 25\% sampled CIFAR100 was used, and the teacher network was set to WResNet40-4. WResNet16-2 and WResNet16-1 were used as student networks that have different dimensional feature maps. Also, we employed MobileNet-V2 which is one of the famous light-weight architecture that has different layer modules, and finally adopted VGG having a quite different architecture with WResNet. Table~\ref{Table3} shows that the proposed method outperforms others for all of the network architectures. For example, in the case of VGG, the proposed method noticeably improved the performance of the student network while the other methods failed or were less effective. This result experimentally proves that the proposed method can distill the network's authentic knowledge that is independent of network architecture.
    
    \begin{table}[t]
        \begin{center}
        \begin{tabular}{c|c|cccc}
        \hline\hline
        Dataset & Student & $\textbf{K}^{int}$ & $\textbf{K}^{alt}$ & IEP & $\textbf{K}^{BB}$\\
        \hline
        \hline
        CIFAR100 & 59.28 & 61.77 & 68.14 & 68.57 & 67.75\\
        TINY & 43.56 & 45.90 & 52.62 & 53.26 & 51.92\\
        \hline
        \end{tabular}
        \caption{Ablation study for each proposed knowledge. TINY denotes TinyImageNet.}\smallskip
        \label{Table4}
        \end{center}
    \end{table}
    
    \begin{table}[t]
        \begin{center}
        \begin{tabular}{c|ccccc}
        \hline\hline
        Sample rate & 100\% & 50\% & 25\% & 10\%\\
        \hline
        \hline
        PCA-IPCA & 78.12 & 74.22 & 68.57 & 55.89\\
        PCA-PCA & 77.92 & 73.66 & 66.58 & 51.88\\
        \hline
        \end{tabular}
        \caption{Performance comparison of PCA and IPCA with various batch sizes.}\smallskip
        \label{Table5}
        \end{center}
    \end{table}
    
    \subsection{Ablation Study}\label{sec.4.6}
        CIFAR100 and TinyImageNet datasets were used for ablation study. For a clear comparison, we experimented with the 25\% sampled datasets showing significant performance improvements. First, Table \ref{Table4} shows how the three types of knowledge, $\textbf{K}^{int}$, $\textbf{K}^{alt}$, and $\textbf{K}^{BB}$, can help improve network performance. It is worth noting that even the student network that has only one knowledge transferred delivers significant performance. In particular, $\textbf{K}^{alt}$ and $\textbf{K}^{BB}$, i.e., knowledge related to alteration, can outperform the SOTA techniques.
        
        Next, we investigated the effect of the SPCA for obtaining the proposed knowledge. As mentioned earlier, if IPCA is not used as the second PCA, accurate PCs will not be obtained, and inter-data relations will not be predicted well. So, MPNN concentrates on memorizing data rather than accurately estimating relations. As a result, network performance is degraded because the intended knowledge is not distilled. Table~\ref{Table5} supports this analysis. When using the normal PCA as the second PCA, the performance tends to decrease, and the lower the sampling rate, the larger the degradation. This is because the distillation module makes it easier to memorize data. Therefore, we can find that IPCA is essential for obtaining the proposed knowledge. For reference, the visualization of the embedded dataset obtained by each method can be found in the supplementary material.
        
        Finally, we take a look at the performance according to the number of iterations of MPNN. As the number of iterations increases, more accurate IEP knowledge can be obtained. However, if MPNN uses too many layers, the capacity increases, so there is a risk of focusing on memorizing data. Table~\ref{Table6} demonstrates this assumption. In fact, we can see that the performance increases as the number of iterations increases, but at a certain point, the performance starts decreasing. Our experiments show that two to three iterations are most advantageous for the best performance.

    \begin{table}[t]
        \begin{center}
        \begin{tabular}{c|ccccc}
        \hline\hline
        Iteration & 1 & 2 & 3 & 4\\
        \hline
        \hline

        CIFAR100     & 67.91 & 68.57 & 68.44 & 66.63\\
        TinyImageNet & 52.81 & 53.20 & 53.28 & 51.55\\

        \hline
        \end{tabular}
        \caption{Performance according to the number of message passing iterations.}\smallskip
        \label{Table6}
        \end{center}
    \end{table}
    
\section{Conclusion}\label{sec.5}
    Knowledge distillation has recently been proven to be effective in a variety of computer vision problems with significant advances. However, research on knowledge distillation is currently too focused on performance. We thought that knowledge distillation could be an essential tool for understanding the nature of deep neural networks, not just a technique. So we validated our idea by defining new interpretable knowledge and suggesting a graph neural network based on interpretable knowledge. Thus, the experimental result shows that the proposed method outperforms SOTA methods in various datasets.

\section{Acknowledgments}
    This work was supported by Institute of Information \& communications Technology Planning \& Evaluation(IITP) grant funded by the Korea government (MSIT) [2020-0-01389, Artificial Intelligence Convergence Research Center (Inha University)] and National R\&D Program through the National Research Foundation of Korea(NRF) funded by Ministry of Science and ICT(2020M3H2A1078119).
    
{
    \bibstyle{aaai21}
    \bibliography{egbib}
}

\section*{Appendix}
\section*{Additional details of the proposed method}
    This section describes the details omitted in the main body of this paper.
    
    \subsection{IPCA}
        We briefly introduce incremental principal component analysis (IPCA), which is a part of the proposed method. Algorithm 1 shows the pseudo code of IPCA. The basic structure is the same as the one introduced in \cite{ipca}, but we replaced the mechanism for finding the mean vector $\textbf{m}^{\textbf{P}}$ with exponential moving average (EMA) for faster update. The decay rate of EMA was set to 0.9.

    \subsection{Training of Message Passing Neural Network}
        Since the proposed message passing neural network (MPNN) is a separate network from the student network, it has a different learning method and different hyper-parameters. Also, since MPNN is rarely dependent on datasets, the training setting of MPNN was the same for all datasets. The iteration number of message passing, \textit{I} was set to 2. \textit{L}$_{2}$-regularization was applied to all trainable variables, and decay rate was set to $5\times10^{-4}$. Learning rate was fixed to 0.1, and stochastic gradient descent (SGD) optimizer \cite{sgd} was used for learning. Also, Nesterov momentum \cite{nesterov} was adopted and momentum was set to 0.9. And regardless of datasets, all networks were trained for 8,000 iterations. In addition, the batch size and data augmentation were matched with those for learning each student network.
        
        \begin{algorithm}[t]
            \SetAlgoNoLine
            \caption{Incremental PCA}\label{alg1}
                $\textbf{Input} : \bar{\textbf{P}}, \textit{is\_Training\_phase}$\\
                $N$, $D$ are $\textbf{P}$'s shape.\\
                \vspace{3pt}
                \textbf{Initialize} :\\
                \hspace{15pt}$\textbf{m}^{\textbf{P}} \leftarrow 0 \in \mathbb{R}^{D}$\\
                \hspace{15pt}$\mathbf{\Sigma}^{\textbf{P}} \leftarrow 0 \in \mathbb{R}^{\frac{D}{2}\times \frac{D}{2}}$\\
                \hspace{15pt}$\textbf{V}^{\textbf{P}} \leftarrow 0 \in \mathbb{R}^{D\times \frac{D}{2}}$\\
                \vspace{3pt}
                \textbf{if} \textit{is\_Training\_phase} \textbf{then}\\
                \hspace{15pt}$\textbf{m}^{c} \leftarrow \frac{1}{N}\sum^{N}_{n=1}\textbf{p}_{n}$\\
                \hspace{15pt}$ \textbf{U}^{\textbf{P}}, \mathbf{\Sigma}^{\textbf{P}}, \textbf{V}^{\textbf{P}*}\leftarrow \text{SVD}\left(\left[\textbf{P}-\textbf{m}_{c},\right.\right.$\\
                \hspace{102pt}$\left.\left.\mathbf{\Sigma}^{\textbf{P}}\cdot \textbf{V}^{\textbf{P}*},\textbf{m}^{\textbf{P}}-\textbf{m}^{\textbf{c}}\right]\right)$\\
                
                \hspace{15pt}$\mathbf{\Sigma}^{\textbf{P}} \leftarrow \left[\sigma^{\textbf{P}}_{v,w} \right]_{1 \leq v,w \leq \frac{D}{2}}$\\
                
                \hspace{15pt}$\textbf{V}^{\textbf{P}} \leftarrow \left[\text{sign}\left(\text{max}\left(\textbf{v}^{\textbf{P}}_{d}\right)\right.\right.$\\
                \hspace{58pt}$\left.\left.+\text{min}\left(\textbf{v}^{\textbf{P}}_{d}\right)   \right)\textbf{v}^{\textbf{P}}_{d}\right]_{1 \leq d \leq \frac{D}{2}}$\\
                
                \hspace{15pt}$\textbf{m}^{\textbf{P}} \leftarrow 0.1\textbf{m}^{\textbf{P}} + 0.9\textbf{m}^{\textbf{c}}$\\
                \textbf{end if}\\
                \vspace{3pt}
                $\textbf{C} \leftarrow \left[\textbf{c}_{d}\right]_{1 \leq d \leq \frac{D}{2}} = \left( \bar{\textbf{P}}-\textbf{m}^{\textbf{P}}\right)\cdot \textbf{v}^{\textbf{P}}_{d}$\\
                \vspace{3pt}
                \textbf{return} $\textbf{C}, \textbf{V}^{\textbf{P}}, \textbf{m}^{\textbf{P}}$
        \end{algorithm}
        
    \subsection*{Multi-Head Graph Distillation}
\begin{figure}[t]\centering
            \includegraphics[width=0.45\textwidth]{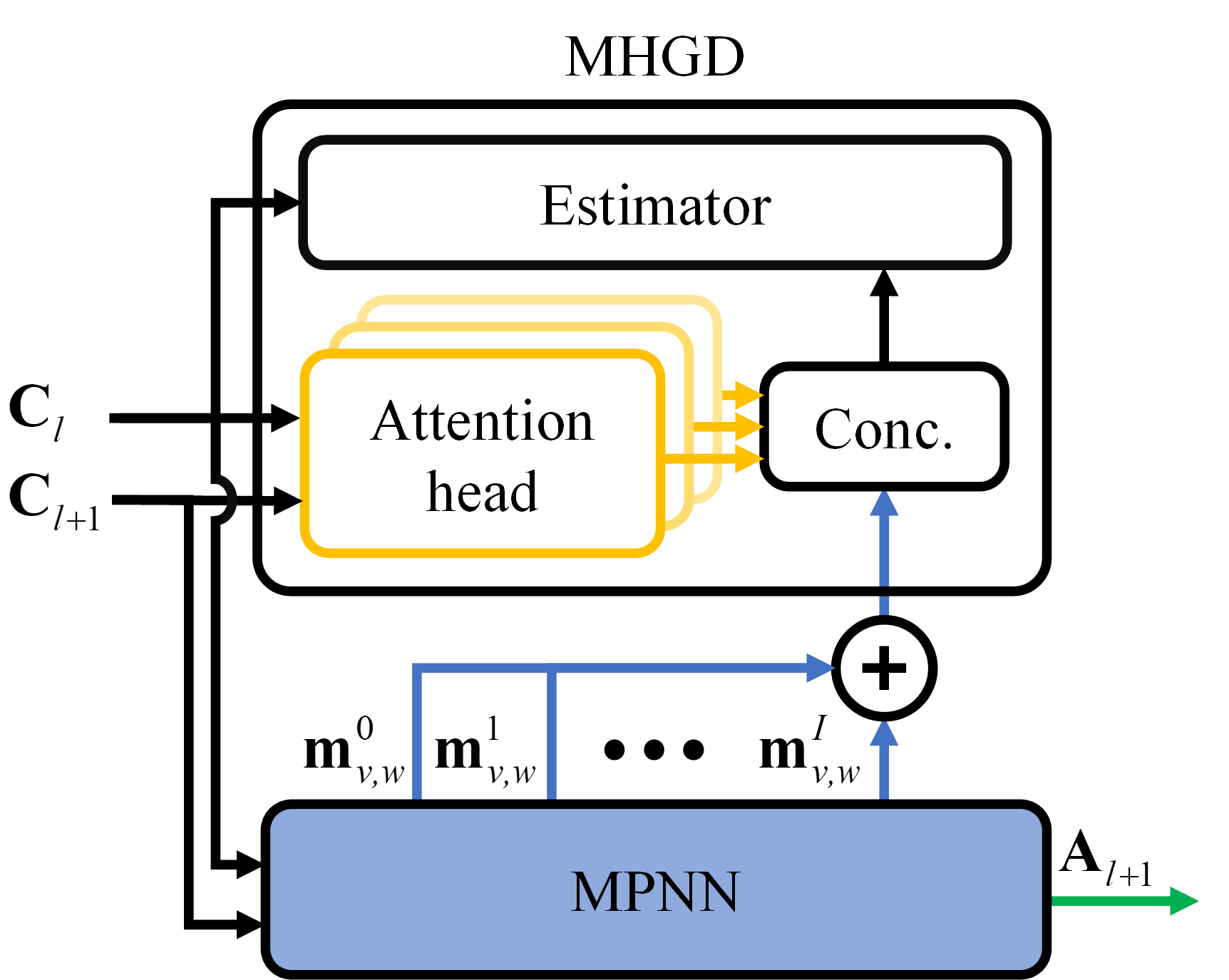}
            \caption{Block diagram of MHGD for distillation of black-box knowledge. Yellow, green, and blue lines represent black-box, interim embedding stages, and alteration of them, respectively.}\label{supp_fig1}
        \end{figure}

        The multi-head graph distillation (MHGD) was implemented as described in the paper \cite{mhgd}. The only difference from \cite{mhgd} is that the message obtained from MPNN was used as extrinsic information for learning the estimator of MHGD. The conceptual diagram is shown in Fig. \ref{supp_fig1}. The number of attention heads was 8, and MHGD was also trained simultaneously with the hyper-parameters of the proposed MPNN. The attention head of MHGD was learned to find the arbitrary relation of $\textbf{C}_{l}$ and $\textbf{C}_{l+1}$. At this time, if the alteration of embedding knowledge $\textbf{K}^{\textit{alt}}$ obtained by MPNN is injected, each attention head will get a relation except $\textbf{K}^{\textit{alt}}$. Since $\textbf{K}^{\textit{alt}}$ is interpretable, the relation obtained by the attention head becomes a non-interpretable relation, or black-box knowledge.
        
    \subsection*{Training Method}
        For fair evaluation, we chose the datasets suitable for each experiment. First, we adopted CIFAR10 \cite{cifar} for visualization of knowledge. Since CIFAR10 has a relatively small number of labels, it is known as an easy-to-learn dataset. Next, for small network enhancement experiment, CIFAR100 \cite{cifar} and TinyImageNet \cite{tinyimagenet} were employed. CIFAR10 and CIFAR100 are identically composed of 50,000 color images of 32$\times$32 pixels, but have 10 and 100 classes, respectively. TinyImageNet consists of 100,000 color images of 64$\times$64 pixels and has 200 classes. Finally, in the experiments to verify transfer learning performance, MIT-scene \cite{mit-scene} and CUB-200-2011 \cite{cub200} datasets were adopted. Note that the two datasets consist of fewer color images compared to the number of classes. The MIT-scene with 67 classes consists of 15,620 images, and the CUB-200-2011 dataset with 200 classes consists of 11,788 images.
    
        Next, random crops and horizontal flipping were applied to CIFAR10 and CIFAR100 \cite{cifar} for augmentation. In the case of MIT-scene \cite{mit-scene}, the training images were randomly resized and cropped at a ratio of [0.75, 1.33] and then resized to 224$\times$224. Each test image was resized to a size of 256$\times$256, and the center portion was cropped to 224$\times$224. For CUB200-2011 \cite{cub200}, augmentation like MIT-scene was applied after cropping to the object's bounding box.
        
        The hyper-parameters for learning networks in each experiment are described as follows. The hyper-parameters used in the other methods were the same as those used in the proposed method. \textit{L}$_{2}$-regularization was applied to all trainable variables, and decay rate was set to $5\times10^{-4}$. The SGD optimizer was used for learning. Also, Nesterov momentum was adopted and momentum was set to 0.9. For the experiment section, the batch size was set to 128, and the initial learning rate starts at 0.1 and decreases by 20\% for 0.3, 0.6, and 0.8 points of learning. For experiments with transfer learning, we used a batch size of 32 and an initial learning rate of 0.01, decreasing 0.1 times at 0.5 and 0.75 points of learning.
        
        Then, look into how to sense feature maps or vectors in the student and teacher networks to transfer knowledge. In the case of activation transfer (AT) \cite{kd-attention}, activation boundary (AB) \cite{AB-KD}, MHGD \cite{mhgd}, and the proposed method, the last of the feature maps with the same depth becomes the sensing point except the stem layer. Thus, three feature maps are obtained for WResNet and four for ResNet. For factor transfer (FT) \cite{kd-ft} and relational knowledge distillation, we used the last feature map and logits vector of each network.
        
\begin{figure}[t]\centering
            \includegraphics[width=0.45\textwidth]{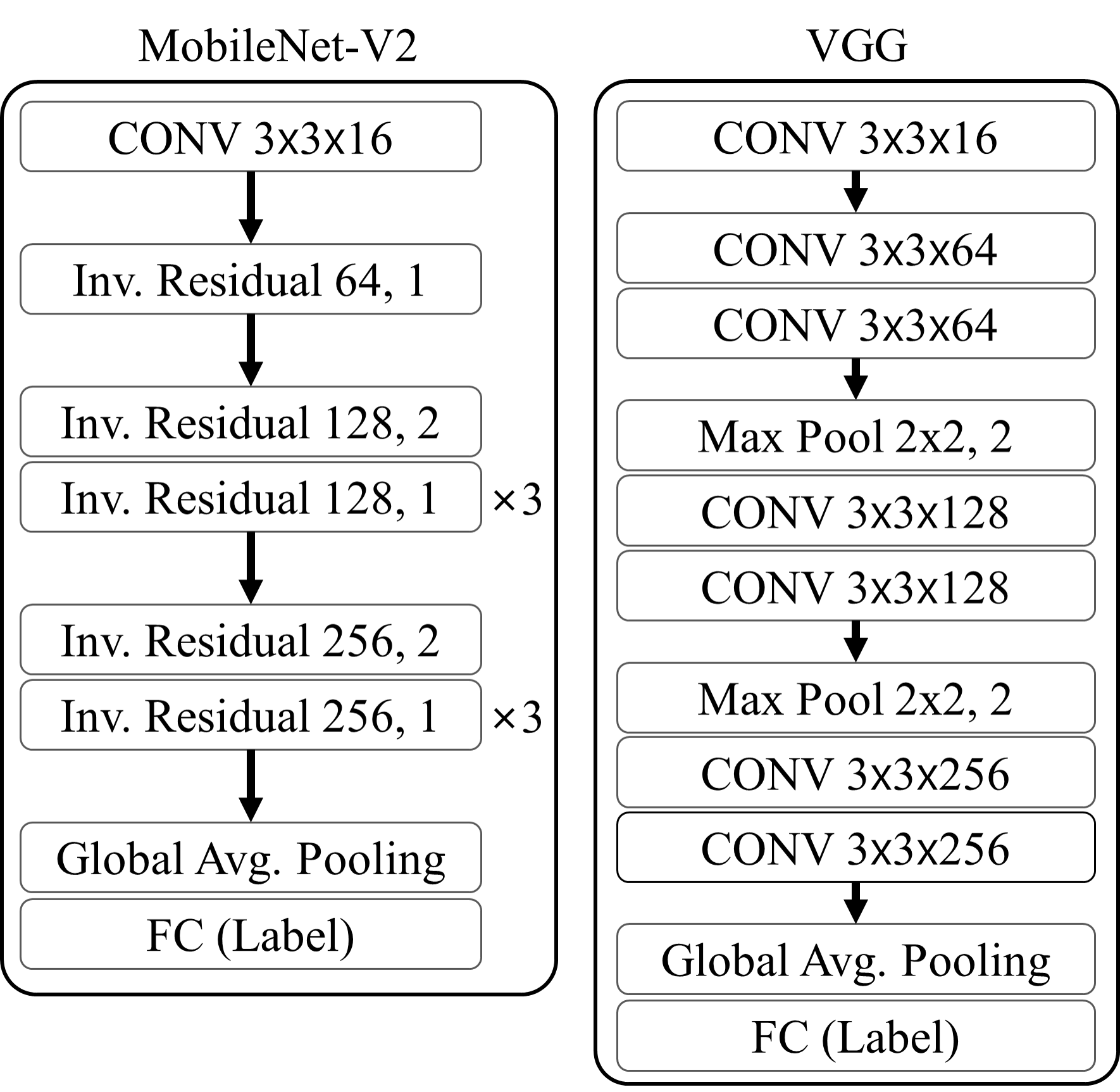}
            \caption{MobileNet-V2 and VGG designed for this experiment. In MobileNet-V2, `Inv. Residual \textit{D}, \textit{s}' represents Inverted Residual block with depth \textit{D} and stride \textit{s}. `Max Pool \textit{r}$\times$\textit{r}, \textit{s}' of VGG indicates max pooling with kernel size of  \textit{r}$\times$\textit{r} and stride \textit{s}.}\label{supp_fig2}
        \end{figure}

        Finally, the network configurations are described. In the case of WResNet \cite{wideresnet} and ResNet \cite{resnet}, we used the model mentioned in the paper. In case of MobileNet-V2 \cite{mobilenetv2} and VGG \cite{vgg} as heterogeneous students, we designed our own models as in Fig. \ref{supp_fig2}. MobileNet-V2 was constructed according to the shape of the teacher network's feature map using the Inverted Residual block as in \cite{mobilenetv2}. VGG was intentionally designed to have a different structure from the teacher network with batch normalization \cite{batchnorm}. Also, leaky ReLU \cite{relu} as the activation function was applied to VGG.
        
\section*{Additional Experimental Results}
    This section introduces some experiments showing the effect of the proposed method more clearly, which were omitted in the main body of this paper.
    \begin{table*}[t]
            \begin{center}
            \begin{tabular}{c|cccccccc}
            \hline\hline
            Acc. & Student & Teacher & SL & FitNet & AT & FSP & DML \\
            \hline\hline
            Last & 71.76 & 78.96 & 71.79 & 72.74 & 72.31 & 72.65 & 73.27\\
            Best & 71.92 & 79.08 & 72.08 & 72.96 & 72.60 & 72.91 & 73.47\\
            \hline\hline
            Acc. & KD-SVD & KD-EID & FT & AB & RKD & MHGD & IEP+BB\\
            \hline\hline
            Last & 73.68 & 73.84 & 73.35 & 73.08 & 73.40 & 73.98 & 74.14\\
            Best & 73.78 & 74.07 & 73.50 & 73.41 & 73.48 & 74.30 & 74.30\\

            \hline
            \end{tabular}
            \caption{Performance comparison of the proposed method and the latest methods. `Acc.' means accuracy, `last' means the last stage of learning, and `best' indicates the best performance among the validation results.}\smallskip
            \label{supp_Table2}
            
            \end{center}
        \end{table*}
    
    \subsection*{Visualization}
        Some experimental results on stacked PCA (SPCA), which can visualize CNN's embedding procedure, were presented in the paper's main body. Here, we visualize the compressed feature map $\textbf{C}$ at each sensing point. For clear visualization, three components were normalized, which are defined by
        \begin{equation}
            \textbf{C}^{\textit{VIS}}_{l}=\left[\frac{c_{n,d}}{\sqrt{\sum^{3}_{k=1} c_{n,k}}} \right]_{1\leq n \leq N, 1\leq d \leq 3}
        \end{equation}
        \begin{equation}
            \textbf{C}_{l}=\left[c_{n,d} \right]_{1\leq n \leq N, 1\leq d \leq 3}
        \end{equation}
        where \textit{N} is the number of samples in dataset, and \textit{l} is the layer to be sensed, $D_{l}$ stands for the dimension of the \textit{l}$^{\textit{th}}$ layer's compressed feature map. The visualization result is given by an attached video file, i.e., IES.mp4. Since $\textbf{C}^{\textit{VIS}}_{0}$ has little progress in embedding, we can see that similar samples are embedded in adjacent positions in terms of low-level features such as color and edge. Green-colored data, which occupies a considerable amount in CIFAR10, was embedded in a very small region, and white-colored data with little information is evenly embedded in a large space. After going through a sufficient number of layers, $\textbf{C}^{\textit{VIS}}_{1}$ starts showing middle-level features. So, the higher-level feature of the green-colored data is reflected and the phenomenon of being embedded in a larger space can be observed. Finally, since $\textbf{C}^{\textit{VIS}}_{2}$ is a sampled feature before the last fully-connected (FC) layer, we can see that it is clustered according to high-level feature, or class. Therefore, SPCA can be used as a tool to interpret the embedding procedure.
    
    \subsection*{Comparison of IPCA and PCA through visualization}
        IPCA can estimate the principal components of a dataset through mini-batch data. On the other hand, since the principal components obtained by a general PCA depend on mini-batch data, the principal components of the exact embedding space cannot be produced. Therefore, feature maps compressed by the ordinary PCA do not account for the embedding space of the dataset. This can be seen in the attached PCA-PCA.mp4. The last feature map is usually clustered according to class, but we can observe that the feature map is randomly distributed because the exact principal components are not available. Thus, the IIR and AIR knowledge obtained by the ordinary PCA do not sufficiently express the embedding procedure knowledge of CNN, and the student network based on IIR and AIR knowledge deteriorates.
        
    \subsection*{Comparison with the State-of-the-Art Methods}
        This section compares the proposed method with the benchmark results of state-of-the-art techniques available at \url{https://github.com/sseung0703/KD_methods_with_TF}. For a fair comparison, we implemented the proposed method based on benchmark codes and trained with the same hyper-parameters. The results of the conventional techniques are the numerical values published in the benchmark results. Table \ref{supp_Table2} shows that the proposed method outperforms conventional methods, i.e SL~\cite{KD}, FitNet~\cite{FitNet}, AT~\cite{kd-attention}, FSP~\cite{FSP}, DML~\cite{dml}, KD-SVD~\cite{KD-SVD}, AB~\cite{AB-KD}, RKD~\cite{rkd}, and MHGD~\cite{mhgd}.

\end{document}